\documentclass[11pt]{article}

\usepackage[final]{acl}

\usepackage{times}
\usepackage{latexsym}

\usepackage[T1]{fontenc}

\usepackage[utf8]{inputenc}

\usepackage{microtype}

\usepackage{inconsolata}

\usepackage{graphicx}
\usepackage{algorithm}
\usepackage{algorithmic}
\usepackage{booktabs} 
\title{TEX: Test-Time Scaling Testing Agents via Execution-based Cross-Validation}

\author{
  \textbf{Akash Gokul\textsuperscript{1*}\textsuperscript{\textdagger}},
  \textbf{Srijan Bansal\textsuperscript{2*}},
  \textbf{Yingbo Zhou\textsuperscript{3}\textsuperscript{\textdagger}},
  \textbf{Semih Yavuz\textsuperscript{2}}
\\
  \textsuperscript{1}NVIDIA,
  \textsuperscript{2}Salesforce Research,
  \textsuperscript{3}Recursive
\\
  \small{
    \textsuperscript{*}Equal contribution.
    \quad \textsuperscript{\textdagger}Work done while at Salesforce Research.
  }
\\
  \small{
    \textbf{Correspondence:} \href{mailto:srijanbansal@salesforce.com}{srijanbansal@salesforce.com}
  }
}

\begin{document}
\maketitle
\begin{abstract}
Test-case generation is critical but remains challenging for language models. Existing test-time scaling methods either isolate candidates or aggregate traces, which do not scale to long-horizon tasks. We present TEX, a collaborative test-time scaling framework based on execution-driven cross-validation. TEX parallelizes agents to generate tests and patches, cross-executing each test against every patch to form a $K{\times}K$ feedback matrix. Agents use their matrix slices for refinement, enabling objective validation without verbose trace sharing. On SWT-Bench Verified, TEX achieves \emph{86.4\%}, establishing a new state-of-the-art and improving +9.0\% over single-agent baselines. Moreover, even random selection from TEX outputs (84.4\%) surpasses the previous SOTA (84.0\%), showing collaborative execution feedback improves solution quality, not just best-case samples. Our results highlight execution-grounded feedback as an effective scaling mechanism for long-horizon agentic tasks.

\end{abstract}

\section{Introduction}

Software engineering agents have made remarkable progress in bug-fixing \citep{jimenez2024swebench, deng2025swebenchproaiagents}, yet \textit{generating high-quality tests} remains challenging. Test-case generation demands both diversity (exploring different test designs) and validation (identifying which tests correctly detect bugs). Current approaches struggle to provide both: isolated generation lacks diversity, while parallel sampling faces a verification gap when no oracle tests exist \citep{li2026benchmarktesttimescalinggeneral}.

Test-time scaling improves large language model (LLM) performance through inference-time computation \citep{openai2026openaio1card,Guo_2025,snell2024scalingllmtesttimecompute}. However, existing methods have limitations. Serial scaling limits diversity and suffers from overthinking \citep{zhou2026thinkinghurtsoverthinkingllm}. Parallel scaling (Best-of-N) requires verifiers \citep{xi2025agentprmprocessrewardmodels,jin2025revealselfevolvingcodeagents}. Hybrid approaches \citep{venkatraman2025recursiveselfaggregation,madaan2025rethinkingthinkingtokensllms} share information across agents but either concatenate full traces (prohibitive at 100K+ tokens) or summarize them (losing details). Moreover, multi-agent collaboration often underperforms single agents under equal budgets \citep{tran2026singleagentllmsoutperformmultiagent}, suggesting inefficient aggregation. 

For test generation, these limitations are acute. Verifier-based methods work when correctness is verifiable, but test quality verification requires oracle code execution. Self-execution frameworks \citep{zheng2025opencodeinterpreterintegratingcodegeneration,ding2024cyclelearningselfrefinecode} run agents' self-generated tests on their own code, providing limited supervision, as failures do not disambiguate between incorrect tests and incorrect code.

\begin{figure}[t]
    \centering
    \includegraphics[width=0.48\textwidth]{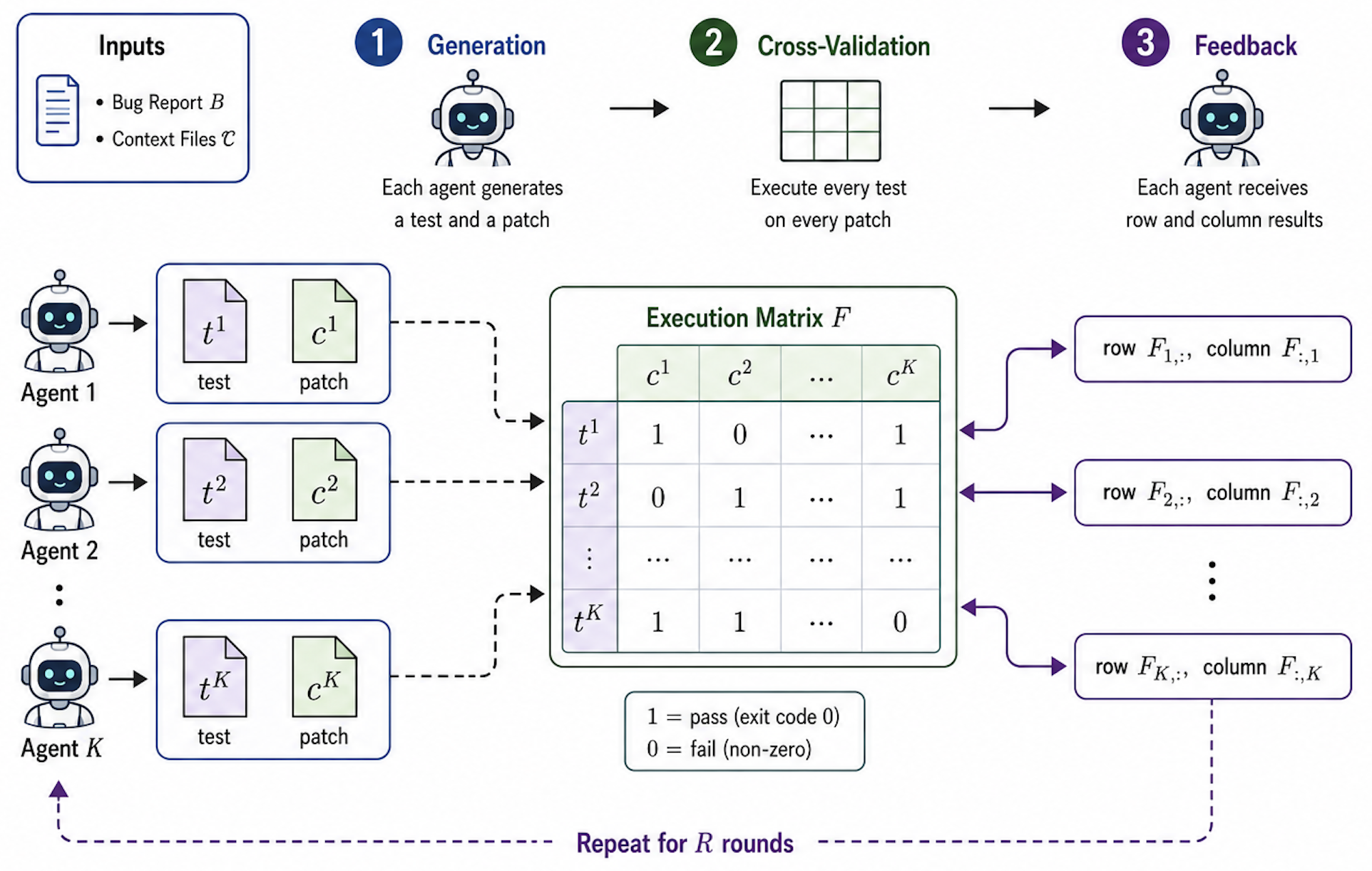}
    \caption{TEX overview. At each round, $K$ agents are sampled in parallel and each agent outputs a test file and code patch. We perform cross-validation by running all generated code patches against all generated test files. Each agent receives execution feedback from running its code \& test against all other tests \& code, then continues sampling in the next round.}
    \label{fig:tex-overview}
\end{figure}

We hypothesize that this can be addressed through \textit{collaborative validation}: agents generate diverse solutions and learn from each other's execution results. We present TEX, which achieves this via \textit{execution-based collaborative learning}. TEX samples $K$ agents in parallel; each generates both test scripts and code patches across multiple rounds. Rather than sharing reasoning traces, we execute every agent's tests against every agent's code, producing a compact $K \times K$ feedback matrix (Figure~\ref{fig:tex-overview}). Each agent receives execution results for cross-candidate refinement. This provides: \textit{(i)} objective verification without learned reward models or oracle verifiers; \textit{(ii)} richer signals than self-execution alone; and \textit{(iii)} compact feedback. TEX achieves \textbf{86.4\%} on SWT-Bench Verified, improving \textbf{+9.0\%} over single-agent baselines while maintaining diversity (Best-of-4 stable at $\sim$89\%). Effective test-time scaling needs the \textit{right signal}: execution feedback, not more context.

\section{Methodology: TEX}

TEX enables collaborative test-case generation through execution-based cross-validation. Rather than sharing reasoning traces, agents share execution results---a grounded signal for long-horizon tasks. Our core hypothesis is that test-case generation benefits from two mechanisms: (1) \textit{active code generation}---agents that write both tests and bug fixes produce better tests than test-only approaches, as understanding the fix helps identify what to test; and (2) \textit{cross-candidate learning}---agents refine their solutions more effectively when they can see how their tests and code patches perform against their peers', rather than in isolation.

\textbf{Method Overview.} Given a bug report, TEX runs $K$ agents in parallel over $R$ rounds (Algorithm~\ref{alg:tex}). Each round proceeds in three phases. \textbf{Generation:} each agent $i$ independently generates test script $T_i$ and code patch $C_i$ (up to $M=8$ steps each)
  based on the bug report, relevant context files, and execution feedback from the previous round ($\mathbf{F}^{r-1}_{i,:}, \mathbf{F}^{r-1}_{:,i}$). Tests aim to
  reproduce the bug (fail on buggy code); patches aim to fix it. \textbf{Cross-Validation:} we execute every agent's test against every agent's code patch,
  building execution feedback matrix $\mathbf{F} \in \{0,1\}^{K \times K}$ where $F_{ij} = 1$ if test $T_i$ passes on code $C_j$ (exit code 0), else $F_{ij} = 0$
  (non-zero exit). This captures all $K^2$ pairwise interactions---does agent $i$'s test correctly validate agent $j$'s fix?---providing multi-source validation
  unavailable in self-execution. \textbf{Feedback:} each agent $i$ receives row $\mathbf{F}_{i,:}$ (how its test performed on all $K$ patches) and column
  $\mathbf{F}_{:,i}$ (how its code performed against all $K$ tests), totaling $2K$ compact execution results. If agent $i$'s test fails on all patches, it may be overly strict; if its code fails most tests, it likely doesn't fix the bug. This signal lets agents distinguish test errors from code errors and refine both in the next round.

\begin{algorithm}[t]
\small
\caption{TEX: Test-Case Generation via Execution Feedback}
\label{alg:tex}
\begin{algorithmic}[1]
\STATE \textbf{Input:} Bug report $B$, context files $\mathcal{C}$, ensemble size $K$, rounds $R$
\FOR{round $r = 1$ to $R$}
    \STATE \textit{// Generation Phase}
    \FOR{agent $i = 1$ to $K$ \textbf{in parallel}}
        \STATE $T_i^r \gets$ GenerateTest($B, \mathcal{C}, \mathbf{F}^{r-1}_{i,:}, \mathbf{F}^{r-1}_{:,i}$)
        \STATE $C_i^r \gets$ GeneratePatch($B, \mathcal{C}, \mathbf{F}^{r-1}_{i,:}, \mathbf{F}^{r-1}_{:,i}$)
    \ENDFOR
    \STATE \textit{// Cross-Validation Phase}
    \FOR{$i, j = 1$ to $K$}
        \STATE $F^r_{ij} \gets$ Execute($T_i^r$, $C_j^r$) \textit{// 1 if pass, 0 if fail}
    \ENDFOR
    \STATE \textit{// Feedback Phase: $\mathbf{F}^r$ provided to agents in next round}
\ENDFOR
\STATE \textit{// Selection Strategy}
\STATE \textbf{Return:} MajorityVote($\{T_1^R, \ldots, T_K^R\}$, $\{C_1^R, \ldots, C_K^R\}$)
\end{algorithmic}
\end{algorithm}

\textbf{Selection Strategy.} After $R$ rounds, we select the final test via majority voting over execution results. We first filter tests that fail to detect the bug (exit code $= 0$ on pre-patch code), then select the one that passes (exit code $= 0$) most frequently across all $K$ generated patches---frequent passing indicates robustness and agreement with multiple independent fixes. If no valid test remains, we randomly select one. This execution-based selection avoids learned verifiers while leveraging code patches as natural validation signals.
\section{Experiments}

\textbf{Setup.} We use the 433-instance subset of SWE-Bench Verified~\citep{openai2024swebenchverified, jimenez2024swebench} introduced by~\citet{mundler2024swtbench} for bug-fixing (TEX-C) and SWT-Bench Verified~\cite{mundler2024swtbench} for test-case generation (TEX-T). This subset excludes instances whose gold test suites are flaky or otherwise unreliable; because such issues originate in the benchmark's gold tests and are often unsolved (as seen in SWE-Bench's GitHub issues), they inject noise into any submission's score. We extend the CodeMonkeys scaffold~\cite{ehrlich2025codemonkeysscalingtesttimecompute} to support cross-candidate execution, using Claude Sonnet 4 (temperature=0.5, $K=4$ agents, $R=3$ rounds, 8 steps per test and code generation respectively). We intentionally use a simple scaffold---no bash, internet, or kernel---to isolate TEX's contribution from tool richness. Context files are from Agentless~\cite{xia2024agentlessdemystifyingllmbasedsoftware}. Baselines: (1) \textit{Single-agent} (R0): prompting without cross-candidate feedback; (2) \textit{Best-of-N}: oracle selection upper bound; (3) \textit{Self-feedback}: agents refine using only their own test/code pairs.

\textbf{SWT-Bench: State-of-the-art test generation.} TEX-T achieves \textbf{86.4\%} on SWT-Bench Verified, establishing a new state-of-the-art and improving +9.0\% over single-agent baselines (Table~\ref{tab:swt-results}). We observe two findings supporting our hypotheses. First, \textbf{active code generation provides substantial gains}: the test-only baseline achieves 71.3\%, while adding code generation improves performance to 77.4\% (+6.1\%). This validates hypothesis 1---agents generating both tests and fixes produce better tests because understanding the fix informs what to test. Second, \textbf{cross-candidate learning provides immediate value}: Pass@1 improves from 77.4\% (R0) to 84.1\% after one round (+6.7\% from cross-candidate feedback alone), then plateaus at 84.4\% by R3. This shows that multi-source validation against diverse peer solutions provides richer signal than isolated refinement. 

Critically, TEX achieves 84.4\% with \textit{random} selection, surpassing the previous SOTA of 84.0\% (L*Agent v1;~\citealp{logicstar2025swtbench}). This shows cross-candidate learning reduces variance and improves quality, not just best-case outputs. Majority voting, selecting tests that pass most frequently across code patches, improves to 86.4\%, using execution results as validation signals \emph{without} LLM judges or learned verifiers. Best-of-4 performance remains stable across rounds (88.7\%$\rightarrow$89.2\%), confirming TEX preserves diversity.

\begin{table}[t]
\centering
\small
\begin{tabular}{lcc}
\toprule
\textbf{Method} & \textbf{Pass@1} & \textbf{Best-of-4} \\
\midrule
\multicolumn{3}{l}{\textit{Baselines}} \\
Test-Only & 71.3\% & -- \\
Test + Code (R0) & 77.4\% & 88.7\% \\
\midrule
\multicolumn{3}{l}{\textit{TEX-T}} \\
Round 1 & 84.1\% & 88.7\% \\
Round 2 & 84.1\% & 88.9\% \\
Round 3 & 84.4\% & 89.2\% \\
\midrule
TEX-T + Majority Vote & \textbf{86.4\%} & -- \\
\midrule
Previous SOTA (L*Agent v1) & 84.0\% & -- \\
\bottomrule
\end{tabular}
\caption{SWT-Bench Verified results.}
\label{tab:swt-results}
\end{table}

\begin{table}[t]
\centering
\small
\begin{tabular}{lcc}
\toprule
\textbf{Method} & \textbf{Pass@1} & \textbf{Best-of-4} \\
\midrule
\multicolumn{3}{l}{\textit{Baseline}} \\
Code Generation (R0) & 57.8\% & 68.4\% \\
\midrule
\multicolumn{3}{l}{\textit{TEX-C}} \\
Round 1 & 58.9\% & 67.7\% \\
Round 2 & 59.4\% & 67.8\% \\
Round 3 & 59.8\% & 67.9\% \\
\bottomrule
\end{tabular}
\caption{SWE-Bench Verified results.}
\label{tab:swe-results}
\end{table}

\textbf{SWE-Bench: Generalization to bug-fixing.} To verify execution-based collaboration generalizes beyond test generation, we evaluate TEX-C on bug-fixing. On SWE-Bench Verified, TEX-C improves Pass@1 from 57.8\% to 59.8\% across three rounds (Table~\ref{tab:swe-results}), showing consistent gains each round (R1: 58.9\%, R2: 59.4\%, R3: 59.8\%). This progression differs from SWT-Bench's plateau after R1, suggesting bug-fixing benefits from sustained refinement while test generation converges faster. This aligns with intuition: fixing bugs requires iterating on complex logic and edge cases, whereas writing tests involves clearer specifications that converge more quickly. The steady improvement validates that cross-candidate feedback continues providing value even when initial gains are smaller. Our baseline (57.8\%) is 7.1\% below mini-swe-agent \citep{yang2024sweagent} (64.9\%) due to our simple scaffold lacking bash, internet, and kernel access. This isolates TEX's gains (+2.0\% absolute) from scaffold improvements, showing cross-candidate learning is independent of tool richness. This isolation is valuable for understanding the method's contribution and suggests future work: combining TEX with more capable scaffolds could reach 66\%--67\% simply by using TEX with mini-swe-agent's harness.

\textbf{Cost-Quality tradeoffs.} TEX exchanges inference compute for solution quality: running $K$ agents for $R$ rounds costs 12$\times$ a single agent (as we use $K=4$, $R=3$) and buys +9.0\%. The exchange compares favorably to Best-of-4 (4$\times$), which reaches 88.7\% only under oracle selection; TEX attains 86.4\% using majority voting over execution signals, recovering most of that gap without LLM judges or learned verifiers. The curve is strongly concave: performance plateaus after Round 1 on SWT-Bench (84.1\%$\rightarrow$84.4\%), so stopping at $R=1$ saves 67\% of the cost for 0.3\% quality loss, and is still well above the single-agent baseline (84.1\% vs.\ 77.4\%).

\section{Analysis}

\subsection{Why Cross-Candidate Learning Works}

\begin{figure}[t]
    \centering
    \includegraphics[width=\linewidth]{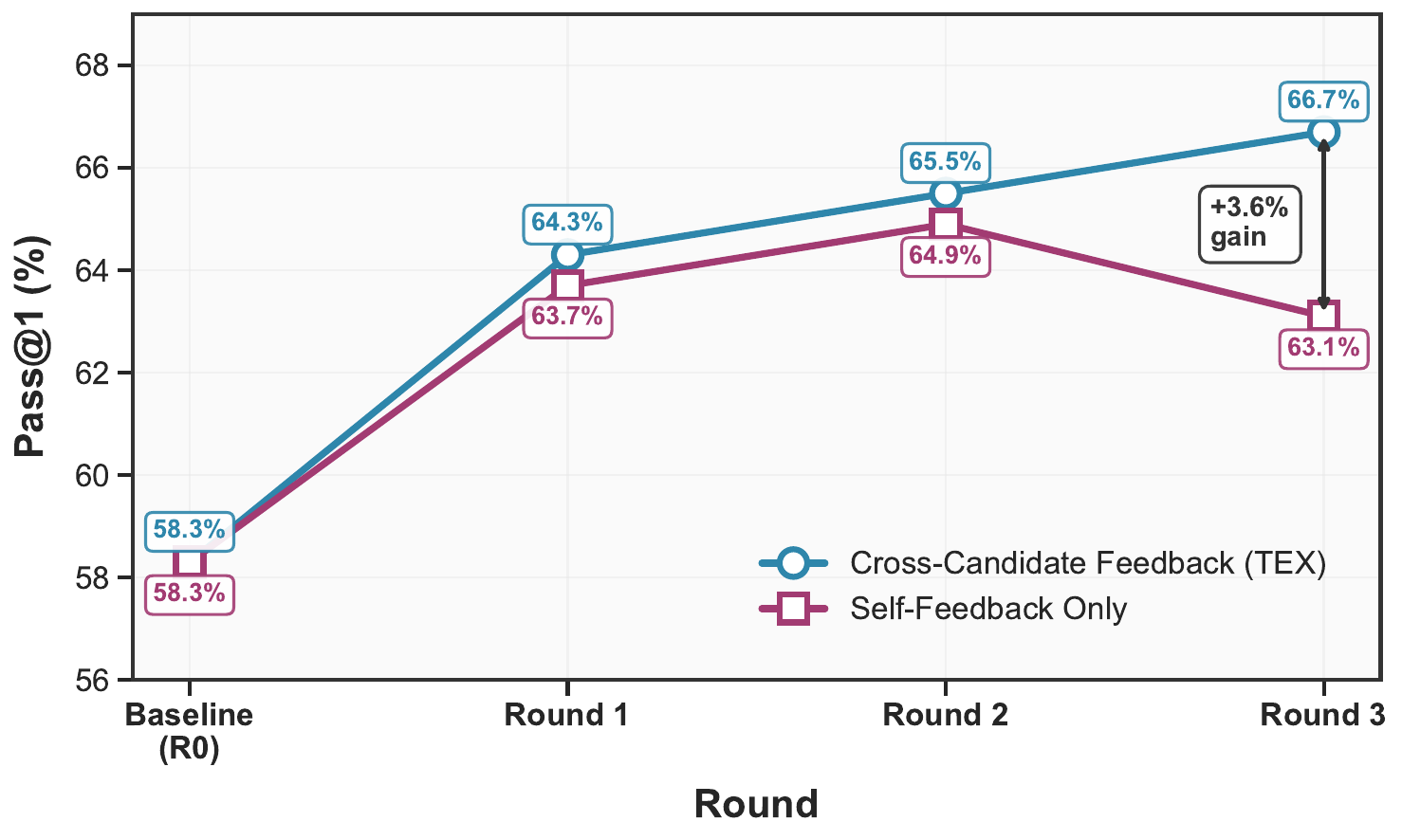}
    \caption{Ablation on SWT-Bench subset shows cross-candidate feedback (TEX) outperforms self-feedback under identical compute ($K=4$, $R=3$).}
    \label{fig:ablation}
\end{figure}

To isolate whether TEX's gains stem from cross-candidate signals rather than extra compute, we compare against \textit{self-feedback}, where agents receive only their own test/code results. Both use identical compute ($K=4$, $R=3$), differing only in feedback scope. Figure~\ref{fig:ablation} shows results on a 42-instance SWT-Bench subset. Self-feedback improves initially but plateaus at Round 2 (64.9\%), then drops to 63.1\% at Round 3, indicating overfitting without external validation. In contrast, cross-candidate feedback sustains gains across rounds (58.3\% at R0 $\rightarrow$ 66.7\% at R3).

\textbf{Three mechanisms explain this advantage.} (1) \textit{Multi-source validation}: when agent $i$'s test fails on agent $j$'s code but passes on agent $k$'s, it learns the test may be overly strict; self-feedback lacks this triangulation and cannot distinguish ``my test is wrong'' from ``my code is wrong.'' (2) \textit{Exposure to alternatives}: cross-candidate feedback avoids local optima by exposing agents to peers' approaches through execution results. (3) \textit{Scalable aggregation}: TEX shares $O(K)$ pass/fail results versus $O(K \cdot L)$ tokens for trace concatenation~\cite{venkatraman2025recursiveselfaggregation,madaan2025rethinkingthinkingtokensllms}, where $L$ often exceeds 100K for software agents.

\textbf{Difficulty stratification explains performance patterns.} Full SWT-Bench plateaus after Round 1 (84.1\%$\rightarrow$84.4\%), while the ablation continues improving. We hypothesize easy instances converge quickly, while harder ones require more rounds. This suggests adaptive stopping: stop at $R=1$ for converged instances (67\% cost reduction), and continue where agents disagree.

\section{Background and Related Work}

\textbf{Test-Time Compute Scaling.} Recent work explores allocating more computation at inference time \citep{snell2024scalingllmtesttimecompute}. Recursive self-aggregation \citep{venkatraman2025recursiveselfaggregation} refines outputs by randomly subsampling and aggregating across parallel samples. Iterative refinement methods \citep{madaan2025rethinkingthinkingtokensllms} improve solutions across multiple rounds. For agentic tasks, General AgentBench \citep{li2026benchmarktesttimescalinggeneral} evaluates test-time scaling, identifying context ceiling and verification gap as bottlenecks. Process reward models \citep{xi2025agentprmprocessrewardmodels} and self-evolving frameworks \citep{jin2025revealselfevolvingcodeagents} address verification through learned value functions.

\textbf{Execution-Based Code Generation.} OpenCodeInterpreter \citep{zheng2025opencodeinterpreterintegratingcodegeneration} and CYCLE \citep{ding2024cyclelearningselfrefinecode} integrate code generation with execution feedback, where models iteratively refine outputs based on test results. These frameworks show execution provides objective validation signals superior to natural language feedback alone. However, they focus on single-agent self-correction rather than multi-agent collaboration.

\textbf{Software Engineering Benchmarks and Agents.} SWE-Bench \citep{jimenez2024swebench} evaluates bug fixing, while SWT-Bench \citep{mundler2024swtbench} evaluates test generation. Prior methods use stronger scaffolds (SWE-Agent \citep{yang2024sweagent}, SWE-Gym \citep{pan2025swegym}) or verifier-based selection ( \citep{shum2025swerm,xi2025agentprmprocessrewardmodels}). TEX instead enables lightweight communication through cross-candidate execution outcomes.

\textbf{Test Generation Methods.} Code-T \citep{chen2022codetcodegenerationgenerated} first demonstrated iterative test generation through execution feedback loops. For SWT-Bench, published work includes e-Otter++ \citep{ahmed2026heterogeneouspromptingexecutionfeedback}, which relies on off-the-shelf code patches to validate generated tests without agents solving the underlying code bug, and L*Agent v1 \citep{logicstar2025swtbench}---a closed-source approach.

\section{Conclusion}

We present TEX, a test-time scaling method achieving \textbf{86.4\%} on SWT-Bench (\textbf{first place}) through collaborative learning via execution feedback. Instead of sharing verbose reasoning traces---impractical for long-horizon software tasks---TEX exchanges compact pass/fail outcomes, enabling scalable iterative refinement for agents generating hundreds of thousands of tokens per instance. Our analysis shows: (1) cross-candidate learning outperforms self-feedback through multi-source validation; (2) collaboration preserves solution diversity rather than causing convergence; and (3) adaptive stopping could reduce cost by 67\% with minimal loss. TEX demonstrates that effective test-time scaling for complex agents requires not more context, but sharing the right structured signals---execution feedback instead of natural language traces.

\section*{Limitations}
TEX's 12$\times$ cost ($K=4$, $R=3$) could be reduced via: (1) adaptive stopping at $R=1$ (67\% reduction, 0.3\% loss), (2) dynamic ensemble sizing by instance difficulty, (3) partial cross-validation (computing a subset of the $K \times K$ matrix). Our simple scaffold isolates TEX's contribution but limits absolute SWE-Bench performance; richer scaffolds (bash, search tools, debugging kernels) could yield larger gains. Our Python-only evaluation requires validation on other languages (Java, C++, JavaScript) and domains beyond GitHub---while execution feedback is language-agnostic in principle, empirical generalization remains future work. We fixed $K=4$ due to budget constraints; systematic exploration of $K \in \{2, 4, 8, 16\}$, together with confidence intervals over repeated runs, could reveal optimal tradeoffs but remains beyond our compute budget; all reported numbers are single-run results.

\noindent\textbf{Potential risks.} TEX amplifies the capabilities of LLM-based test generation agents, which raises concerns shared broadly by autonomous test-case generation systems: generated tests may be overly permissive (missing critical edge cases or security vulnerabilities) or overly strict (rejecting valid implementations), and these issues may be difficult to detect without domain expertise. Because TEX employs cross-candidate learning where agents observe each other's execution results, there is a risk that systematic biases or blind spots shared across the agent ensemble could be reinforced rather than corrected---if all agents miss the same class of bugs, cross-validation provides no corrective signal. Additionally, agreement-based selection could amplify common failure modes if most agents converge to similar but flawed test designs. We recommend that TEX-generated tests undergo the same review and validation processes applied to human-authored tests, including manual inspection for completeness, correctness, and security considerations before deployment in production systems.

\section*{Acknowledgments}
We would like to thank Hiroaki Hayashi and Wenting Zhao for useful discussions. We would also like to thank Srinath Reddy for infrastructure support. The authors used LLM-based tools (Claude Code) to assist with manuscript editing, including rewording, restructuring sections, condensing verbose text, and typographical corrections. All scientific content, experimental design, implementation, analysis, and interpretation were conducted by the authors.

\bibliography{custom}

\end{document}